%% file: main_eng.tex
\documentclass[preprint,12pt]{elsarticle}
\usepackage{natbib}
\usepackage{amssymb}
\usepackage{amsmath}
\usepackage{graphicx}
\usepackage{algorithm}
\usepackage{algpseudocode}
\usepackage{array}   
\usepackage{booktabs} 
\usepackage{microtype}
\biboptions{authoryear}

\journal{Information Processing \& Management}

\begin{document}

\begin{frontmatter}

\title{Argument Rarity-based Originality Assessment for AI-Assisted Writing}

\author[1,2]{Keito Inoshita\corref{cor1}}
\ead{inosita.2865@gmail.com}
\author[3]{Michiaki Omura}
\ead{omura@arbege.com}
\author[4]{Tsukasa Yamanaka}
\ead{yaman@fc.ritsumei.ac.jp}
\author[5]{Go Maeda}
\ead{maeda-g@st.ritsumei.ac.jp}
\author[5]{Kentaro Tsuji}
\ead{k-tsuji@st.ritsumei.ac.jp}

\cortext[cor1]{Corresponding author.}

\affiliation[1]{organization={Ritsumeikan Global Innovation Research Organization (R-GIRO), Ritsumeikan University},
            addressline={1-1-1, Nojihigashi},
            city={Kusatsu},
            postcode={525-8577},
            state={Shiga},
            country={Japan}}

\affiliation[2]{organization={Faculty of Business and Commerce, Kansai University},
            addressline={3-3-35, Yamatecho}, 
            city={Suita},
            postcode={564-8680}, 
            state={Osaka},
            country={Japan}}

\affiliation[3]{organization={Arbege Corporation},
            addressline={3-20, Yotsuyatori}, 
            city={Nagoya},
            postcode={464-0819}, 
            state={Aichi},
            country={Japan}}

\affiliation[4]{organization={College of Life Sciences, Ritsumeikan University},
            addressline={1-1-1, Nojihigashi},
            city={Kusatsu},
            postcode={525-8577},
            state={Shiga},
            country={Japan}}

\affiliation[5]{organization={Office of General Education, Ritsumeikan University},
            addressline={1-1-1, Nojihigashi},
            city={Kusatsu},
            postcode={525-8577},
            state={Shiga},
            country={Japan}}

\begin{abstract}
This study proposes Argument Rarity-based Originality Assessment (AROA), a framework for automatically evaluating argumentative originality in student essays. AROA defines originality as rarity within a reference corpus and evaluates it through four complementary components: structural rarity, claim rarity, evidence rarity, and cognitive depth, quantified via density estimation and integrated with quality adjustment. Experiments using 1,375 human essays and 1,000 AI-generated essays on two argumentative topics revealed three key findings. First, a strong negative correlation ($r = -0.67$) between text quality and claim rarity demonstrates a quality-originality trade-off. Second, while AI essays achieved near-perfect quality scores ($Q = 0.998$), their claim rarity was approximately one-fifth of human levels (AI: 0.037, human: 0.170), indicating that LLMs can reproduce argumentative structure but not semantic originality. Third, the four components showed low mutual correlations ($r = 0.06$--$0.13$ between structural and semantic dimensions), confirming that they capture genuinely independent aspects of originality. These results suggest that writing assessment in the AI era must shift from quality to originality.
\end{abstract}




\end{frontmatter}

\input{Authors/documents/1_introduction}
\input{Authors/documents/2_related-work}
\input{Authors/documents/3_methodology}
\input{Authors/documents/4_experiment-and-analysis}
\input{Authors/documents/5_discussion}
\input{Authors/documents/6_conclusion}

\section*{Declaration of competing interest}
The authors declare that they have no known competing financial interests or personal relationships that could have appeared to influence the work reported in this paper.

\section*{Acknowledgment}
This work was supported by a research grant from the Ritsumeikan Global Innovation Research Organization (R-GIRO), Ritsumeikan University. The authors also acknowledge the RaaS (Ritsumeikan AI-powered Assessment Solution) system developed and deployed at Ritsumeikan University, which provided practical insights into large-scale AI-assisted assessment (https://raas.ritsumei.ac.jp/).

\section*{Data availability}
The AIDE dataset used in this study is publicly available at Kaggle (https://www.kaggle.com/datasets/lburleigh/tla-lab-ai-detection-for-essays-aide-dataset). The code for AROA will be made available upon request.

\section*{Declaration of Generative AI Use}
During the preparation of this manuscript, the authors used generative AI tools for language editing and organization. All content was reviewed and verified by the authors, who take full responsibility for the final manuscript.

\bibliographystyle{elsarticle-harv}
\bibliography{refs}

\end{document}

%% file: Authors/documents/1_introduction.tex
\section{Introduction}
\label{sec:introduction}
\subsection{Background and Motivation}
\label{sec:background}

The rapid advancement of Large Language Models (LLMs) is fundamentally transforming human intellectual activities \citep{1,2}. LLMs such as ChatGPT can instantly generate logically coherent, fluent, and persuasive text on any topic. Writing well is no longer an exclusively human capability, which has a particularly profound impact on education, where determining whether a student's essay reflects original thinking or merely reproduces typical patterns has become increasingly difficult.

Historically, writing assessment has emphasized quality, logical coherence, grammatical accuracy, organizational clarity, and persuasive argumentation. Automated Essay Scoring (AES) research has primarily focused on predicting these quality indicators \citep{3}. However, now that LLMs can surpass humans in text quality, such criteria alone are no longer sufficient indicators of human thinking ability or learning outcomes.

The essential purpose of having students write is not the production of well-formed text itself but the cultivation of critical thinking: forming original perspectives, exploring diverse evidence, and anticipating counterarguments \citep{4}. From this perspective, the paradigm of assessment must shift from quality to originality.

This study conceptualizes argumentative originality as four mutually independent dimensions: i) the uniqueness of argument structure: whether the pattern of claims, reasons, evidence, and counterarguments is typical or rare; ii) the uniqueness of claim content: whether the central assertion represents an original perspective; iii) the uniqueness of evidence selection: whether the argument employs rare, independently explored evidence; and iv) the depth of reasoning: whether the argument develops multi-layered reasoning with counterargument-rebuttal exchanges. These dimensions are mutually independent, and comprehensive originality assessment requires capturing each individually.

However, consistent evaluation of such originality by human raters is inherently difficult. Instructors face order effects and fatigue when grading large numbers of essays \citep{5}, their own expertise introduces bias \citep{6}, and judging relative rarity across an entire corpus exceeds human cognitive capacity. LLMs offer a promising alternative: they are immune to fatigue and order effects, can evaluate essays according to uniform criteria, and can process large-scale corpora to objectively calculate relative positioning \citep{7,8}. Indeed, AI-assisted assessment is already being implemented in practice; for example, the RaaS system deployed at a large Japanese university supports courses with up to 1,000 students per class \citep{36}. Despite such implementations, most existing AI-assisted assessment research has focused on quality evaluation; no framework has been established for systematically evaluating originality, which is an essentially different concept from quality and requires its own theoretical foundation.

This study proposes Argument Rarity-based Originality Assessment (AROA), a framework for automatically evaluating argumentative originality. The core idea of AROA is to operationalize originality as rarity within a reference corpus and to measure it quantitatively using density estimation. AROA evaluates originality through four complementary components: structural rarity (atypicality of argument structure via kernel density estimation), claim rarity (semantic distance of central claims via neighborhood density of sentence embeddings), evidence rarity (uniqueness of supporting evidence using the same methodology), and cognitive depth (dialogic structure and hierarchical depth of reasoning). These components have been confirmed through correlation analysis to have low mutual correlations, each capturing different aspects of originality. AROA integrates them after adjusting with quality scores, thereby treating quality and originality as independent evaluation axes.

\subsection{Research Objectives}
\label{sec:research_objectives}

Based on the above background and motivation, this study addresses the following three research questions:

\begin{enumerate}
    \item[RQ1:] Can argumentative originality in essays be automatically evaluated in a multifaceted manner through density-based rarity quantification of argument components?
    \item[RQ2:] What is the relationship between text quality and argumentative originality, and does a trade-off exist between them?
    \item[RQ3:] How do the originality profiles of human essays and AI-generated essays differ across structural and semantic dimensions?
\end{enumerate}

To address these research questions, this study proposes the AROA framework and conducts large-scale experiments using 1,375 human essays and 1,000 AI-generated essays. The contributions are summarized as follows:

\begin{enumerate}
    \item[i)] AROA is a novel framework that automatically evaluates argumentative originality through four complementary components---structural rarity, claim rarity, evidence rarity, and cognitive depth---integrating density-based rarity quantification with quality adjustment.
    
    \item[ii)] Empirical demonstration of a strong negative correlation ($r = -0.67$) between text quality and claim rarity through large-scale experiments, revealing a quality-originality trade-off where logically coherent, high-quality texts tend to rely on typical claim patterns and common evidence.
    
    \item[iii)] Comparative analysis revealing that contemporary LLMs achieve comparable structural complexity to humans in argumentation while exhibiting substantially lower semantic originality of claims (approximately one-fifth of human levels), thereby presenting new guidelines for originality assessment in the AI era.
\end{enumerate}

The remainder of this paper is organized as follows. Section 2 reviews related work and clarifies the positioning of this study. Section 3 describes the proposed framework AROA. Section 4 presents experiments and results. Section 5 discusses implications and limitations. Finally, Section 6 presents the conclusions.

%% file: Authors/documents/2_related-work.tex
\section{Related Work}
\label{sec:related_work}

\subsection{Automated Essay Scoring}
\label{subsec:aes}

AES is a technology that automatically assigns scores to essays using computers. From the late 1990s to the 2000s, systems such as e-rater~\citep{9}, IntelliMetric, and Intelligent Essay Assessor were deployed in large-scale standardized tests. Early systems extracted handcrafted features such as lexical diversity and syntactic complexity, and predicted scores using regression models. However,~\citet{10} demonstrated that content-poor texts satisfying formal features could receive high scores, revealing that early AES relied heavily on surface features.

With the advancement of deep learning since 2016, the technical paradigm shifted. \citet{11} proposed an end-to-end AES using LSTM, demonstrating the effectiveness of neural approaches. \citet{12} systematically classified AES research and identified three major challenges: human-likeness, deception resistance, and creativity assessment.

More recently, LLMs have been applied to AES. \citet{7} evaluated discourse coherence using GPT-4, demonstrating the possibility of explainable AES. \citet{13} showed that novice evaluators referring to LLM feedback achieved accuracy close to experts, redefining AES as human capability augmentation rather than complete automation. However, \citet{14} reported that LLM-based AES tends to rate AI-generated texts higher than human texts, revealing a fundamental bias problem. \citet{38} demonstrated that LLMs can serve as effective tools for early prediction of student performance, further expanding the role of LLMs in educational assessment.

Throughout this evolution, AES research has consistently focused on predicting text quality rather than evaluating originality of argumentation. As \citet{15} noted, creativity and critical thinking assessment remain recognized but unaddressed challenges. AROA addresses this gap by introducing argumentative originality as an evaluation axis independent of quality.

\subsection{Argumentation Mining}
\label{subsec:argumentation_mining}

Argumentation mining automatically identifies and extracts argument structures from natural language text. \citet{16} provided a comprehensive survey covering claim identification, support and rebuttal relation extraction, and mapping to argumentation schemes. The Toulmin model~\citep{17}, which structures argumentation through six elements (Claim, Grounds, Warrant, Backing, Qualifier, and Rebuttal), is widely adopted as the theoretical foundation for argumentative essay analysis.

\citet{18} initiated argumentation mining research targeting student essays by proposing an annotation scheme for claims, premises, and support/attack relations. \citet{19} extended this direction from structure extraction to quality assessment by evaluating argument organization and strength. In educational applications, \citet{20} integrated argument features into AES and demonstrated improved accuracy in persuasive essay scoring, while \citet{21} developed a system that visualizes argument structures and provides feedback to promote learners' self-correction.

Recent LLM-based approaches have further advanced the field. \citet{22} proposed a multi-task benchmark revealing that LLMs show significant performance degradation when performing chained argument understanding. \citet{37} provided a comprehensive survey of computational argumentation, organizing the field into argumentation mining, assessment, and generation, yet originality assessment of argument content remains outside the scope of existing frameworks.

However, existing research evaluates argument quality (structural completeness, organization, persuasiveness) rather than originality of argument content. AROA utilizes argumentation mining techniques to extract argument elements and then quantifies their semantic rarity through density estimation, presenting a direction of evaluating originality rather than quality.

\subsection{Originality and Creativity Assessment}
\label{subsec:originality}

Originality and creativity assessment has a long tradition in psychology and education. \citet{23} defined creativity as the combination of novelty and usefulness, with divergent thinking tests, particularly the Alternative Uses Test (AUT), serving as the representative measurement paradigm.

Traditional originality assessment relied on human raters subjectively judging response rarity, but automatic evaluation based on semantic distance has since become mainstream. \citet{24} proposed evaluating originality by calculating semantic distance using sentence embeddings, establishing the standard approach for automatic creativity assessment. \citet{25} verified the multilingual validity of this approach across 12 languages and over 6,500 participants.

LLM-based creativity assessment has also advanced. \citet{26} fine-tuned LLMs for metaphor generation and achieved agreement with human evaluation exceeding semantic distance methods. \citet{27} reviewed automatic scoring methods for divergent thinking tests, noting that while LLMs are approaching human-level evaluation, the opacity of judgment rationales remains a challenge. \citet{28} analyzed AI-generated divergent thinking responses, showing that AI generates numerous but often irrelevant and redundant ideas.

These studies have succeeded in evaluating novelty of short responses in divergent thinking tests. However, no research evaluates argumentative originality in long texts such as essays, specifically how rare argument elements (claims, evidence, counterarguments) are relative to a reference corpus. AROA extends the concept of rarity as originality, established in divergent thinking research, to argumentative essays by integrating rarity-based evaluation with argumentation mining across four complementary components.

%% file: Authors/documents/3_methodology.tex
\section{Argument Rarity-based Originality Assessment}
\label{sec:methodology}

\subsection{Framework Overview}
\label{subsec:overview}

\begin{figure}[t]
    \centering
    \includegraphics[width=\textwidth]{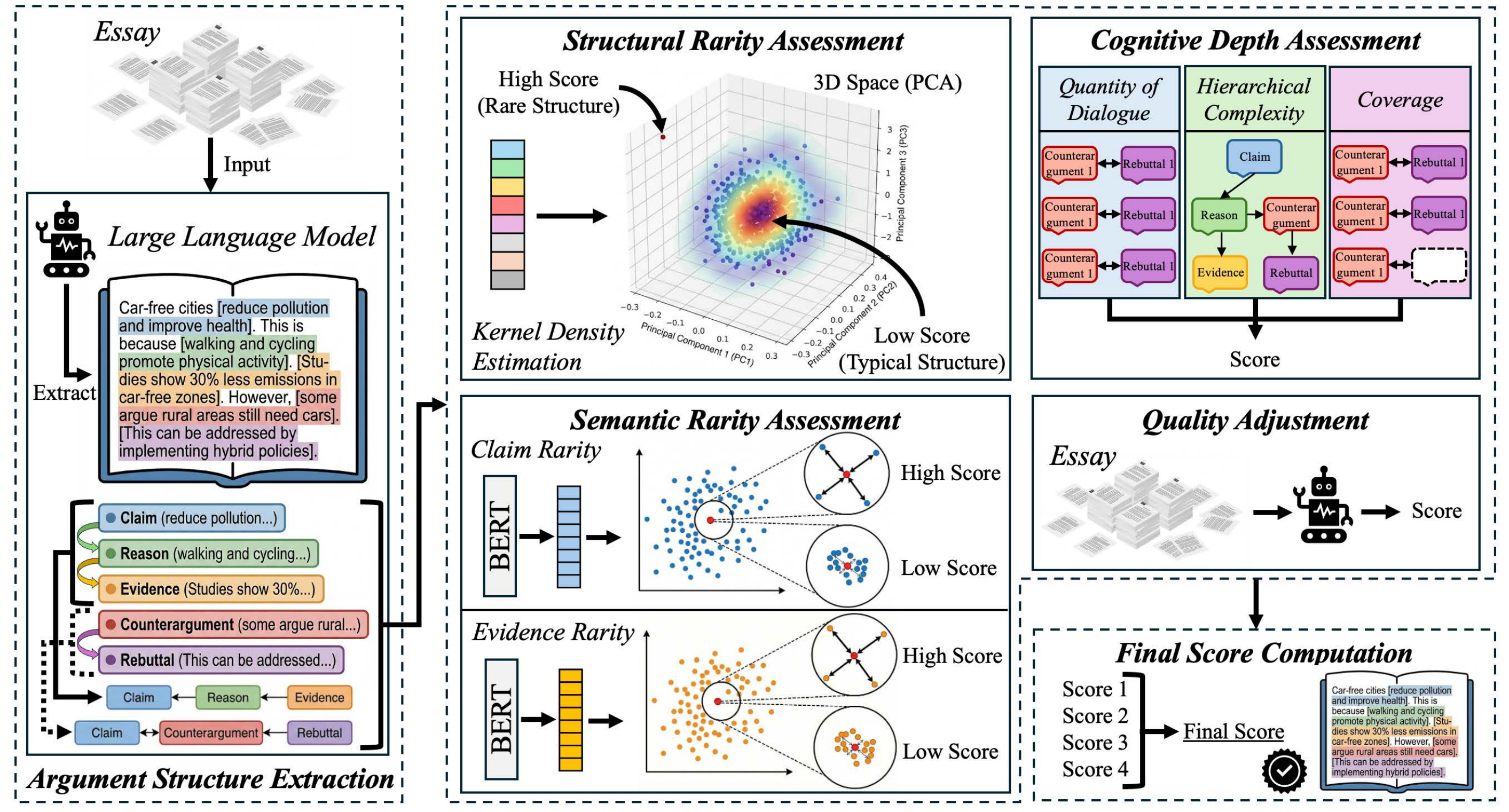}
    \caption{Overview of the AROA processing pipeline.}
    \label{fig:framework}
\end{figure}

AROA is a structure-oriented originality assessment framework that evaluates essays based on the originality of thinking rather than surface fluency or author attribution. Fig.~\ref{fig:framework} shows the overall processing pipeline of AROA. The design philosophy of AROA is based on Toulmin's argumentation model~\citep{17} and integrates insights from critical thinking education~\citep{29, 30}, deep learning theory~\citep{31}, and knowledge construction theory~\citep{32}.

The AROA processing pipeline consists of five stages. First, given a collection of essays on the same topic $\mathcal{D} = \{x_i\}_{i=1}^N$, each essay is transformed using an LLM into a semantic structure $\mathcal{S}(x)$ consisting of five elements: Claim, Reason, Evidence, Counterargument, and Rebuttal. Next, each extracted element is converted into a $d$-dimensional embedding vector using Sentence-BERT, enabling quantitative computation of semantic similarity between elements. Subsequently, based on the embedding representations and structural information, originality sub-scores are calculated from the following four perspectives:
\begin{enumerate}
    \item[i)] Structural rarity ($S_{\text{struct}}$): Evaluates the atypicality of reasoning structure through density estimation
    \item[ii)] Claim rarity ($S_{\text{claim}}$): Evaluates the semantic rarity of central claims using the $K$-nearest neighbor method
    \item[iii)] Evidence rarity ($S_{\text{evid}}$): Evaluates the use of specific and rare evidence
    \item[iv)] Cognitive depth ($S_{\text{move}}$): Evaluates the correspondence between counterarguments and rebuttals, and the depth of reasoning paths
\end{enumerate}
Furthermore, to prevent essays that are original but logically flawed from receiving high scores, a quality score $Q(x)$ that does not depend on fluency is calculated. Finally, the four originality sub-scores are z-score normalized and then integrated through weighted combination, multiplied by the quality score to compute the final evaluation value $S_{\text{final}}(x)$.

These four perspectives capture different dimensions of originality: how arguments are made (structure), what is argued (claims and evidence), and how deeply arguments are developed (cognitive depth). The following sections detail each processing stage.

\subsection{Argument Structure Extraction}
\label{subsec:structure_extraction}

For each essay $x$, we define a structural representation consisting of the following five elements:
\begin{equation}
    \mathcal{S}(x) = \bigl\{ C(x),\; R(x),\; E(x),\; A(x),\; B(x) \bigr\}
    \label{eq:structure}
\end{equation}
The definition of each element is shown in Table~\ref{tab:five_elements}. This five-element representation adapts Toulmin's model~\citep{17} by integrating Warrant and Backing into Reasons and omitting Qualifier, improving practicality in educational settings and consistency of LLM extraction.

\begin{table}[t]
    \centering
    \caption{Definition of five argument elements}
    \label{tab:five_elements}
    \begin{tabular}{clp{6.5cm}}
        \hline
        Symbol & Name & Description \\
        \hline
        $C(x) = \{c_k\}_{k=1}^{n_C}$ & Claims & Central conclusions or assertions that the author wants the reader to accept \\
        $R(x) = \{r_k\}_{k=1}^{n_R}$ & Reasons & Arguments that support the claims; fundamental grounds for why the claims should be accepted \\
        $E(x) = \{e_k\}_{k=1}^{n_E}$ & Evidences & Examples, data, statistics, or citations that back up the reasons \\
        $A(x) = \{a_k\}_{k=1}^{n_A}$ & Counterarguments & Anticipated opposing views, criticisms, or alternative interpretations \\
        $B(x) = \{b_k\}_{k=1}^{n_B}$ & Rebuttals & Responses to counterarguments; refutations or re-assertions \\
        \hline
    \end{tabular}
\end{table}

These elements form hierarchical relationships: Reasons support Claims, Evidence backs up Reasons, Counterarguments oppose Claims, and Rebuttals respond to Counterarguments. This structure captures not only element presence but also the logical composition of argumentation.

An LLM is used for structure extraction with a system prompt specifying the task definition and guidelines:
\begin{quote}
\textit{You are an expert in argument structure analysis. Extract the following five elements from the given essay: (1) Claims: Central conclusions or assertions of the essay, (2) Reasons: Arguments that support the claims, (3) Evidences: Examples, data, or specific information that back up the reasons, (4) Counterarguments: Anticipated opposing views or alternative interpretations, (5) Rebuttals: Responses to counterarguments, re-assertions. Also identify relationships between elements: supports (reason supports claim, evidence backs reason), opposes (counterargument opposes claim), responds\_to (rebuttal responds to counterargument). Guidelines: Preserve the original meaning; avoid excessive summarization. Split compound sentences into semantic units. If an element does not exist, use an empty array.}
\end{quote}
These guidelines ensure faithful extraction, unified element granularity, and handling of missing elements. The essay text is provided in the user prompt with instructions for JSON-formatted output including element lists and relational information. The temperature parameter was set to $T = 0$ for deterministic output.

From the extracted structure, two derived features are calculated. The depth of reasoning $\text{Depth}(x)$ represents hierarchical complexity:

\begin{equation}
\text{Depth}(x) =
\begin{cases}
1, & \parbox[t]{7cm}{if only claims exist} \\
2, & \parbox[t]{7cm}{if reasons exist} \\
3, & \parbox[t]{7cm}{if evidence exists, or both counterarguments and rebuttals exist}
\end{cases}
\end{equation}

This definition distinguishes simple argumentation (depth 1) from complex argumentation with evidence or counterargument-rebuttal structure (depth 3). The number of counterargument-rebuttal matched pairs $n_{\text{matched}}$ is counted from responds\_to relationships, or estimated as $n_{\text{matched}} = \min(n_A, n_B)$ when relationships are not explicitly extracted.

Each obtained text element is projected into vector space by the sentence embedding function $f: \text{Text} \to \mathbb{R}^d$:
\begin{equation}
    \mathbf{v}(c_k),\; \mathbf{v}(r_k),\; \mathbf{v}(e_k),\; \mathbf{v}(a_k),\; \mathbf{v}(b_k) \in \mathbb{R}^d
    \label{eq:embeddings}
\end{equation}
This study used a pre-trained Sentence-BERT model~\citep{33} to obtain $d = 768$-dimensional embeddings. These representations serve as the basis for semantic rarity assessment in subsequent sections.

\subsection{Structural Rarity Assessment}
\label{subsec:structural_rarity}

The structural rarity score $S_{\text{struct}}$ quantifies how rare an essay's argument structure is within the corpus. A 12-dimensional structural feature vector $\boldsymbol{\phi}(x)$ is constructed from each essay $x$:
\begin{equation}
    \boldsymbol{\phi}(x) = \begin{pmatrix}
        n_C \\
        n_R \\
        n_E \\
        n_A \\
        n_B \\
        n_R / \max(n_C, 1) \\
        n_E / \max(n_R, 1) \\
        \text{Depth}(x) \\
        \text{Branches}(x) \\
        \mathbb{I}_{n_A > 0} \\
        \mathbb{I}_{n_B > 0} \\
        n_{\text{matched}}
    \end{pmatrix}
    \label{eq:structural_features}
\end{equation}
Here, $n_C, n_R, n_E, n_A, n_B$ are the counts of each element, $n_R / \max(n_C, 1)$ is the number of reasons per claim, and $n_E / \max(n_R, 1)$ is the number of evidences per reason. $\text{Depth}(x)$ and $n_{\text{matched}}$ are the depth of reasoning and the number of counterargument-rebuttal matched pairs defined in the previous section. $\text{Branches}(x) = \max(n_C, 1) \times \max(n_R, 1)$ represents the number of argument branches, capturing multifaceted argument structures that support multiple claims with multiple reasons. $\mathbb{I}_{n_A > 0}$ and $\mathbb{I}_{n_B > 0}$ are indicator functions showing the presence or absence of counterarguments and rebuttals. These features capture quantitative (element counts), structural (ratios, depth, branches), and critical thinking aspects (counterargument-rebuttal correspondence).

The structural rarity score is defined as the negative log-density in the feature vector space, corresponding to self-information in information theory. Prior to density estimation, standardization is performed on the set of feature vectors across the entire corpus $\Phi = \{\boldsymbol{\phi}(x_i)\}_{i=1}^N$:
\begin{equation}
    \boldsymbol{\phi}'(x) = \frac{\boldsymbol{\phi}(x) - \boldsymbol{\mu}_\Phi}{\boldsymbol{\sigma}_\Phi}
    \label{eq:standardization}
\end{equation}
Here, $\boldsymbol{\mu}_\Phi$ and $\boldsymbol{\sigma}_\Phi$ are the corpus-wide mean and standard deviation of each feature. Dimensionality is then reduced to 3 using PCA to mitigate the curse of dimensionality, and Kernel Density Estimation (KDE) with Scott's rule bandwidth is applied:
\begin{equation}
    S_{\text{struct}}(x) = -\log \hat{p}(\boldsymbol{\phi}'(x))
    \label{eq:s_struct}
\end{equation}
Here, $\hat{p}(\boldsymbol{\phi}'(x))$ is the KDE density estimate after standardization and PCA. Essays with typical structures receive low scores, while those with rare structures receive high scores.

\subsection{Semantic Rarity Assessment}
\label{subsec:semantic_rarity}

The semantic rarity scores $S_{\text{claim}}$ and $S_{\text{evid}}$ quantify how rare the content of claims and evidence are within the corpus. While structural rarity evaluates how arguments are made, semantic rarity evaluates what is argued.

\subsubsection{Claim Rarity}
Claim rarity is evaluated based on how isolated each claim's embedding vector is in semantic space. Let the set of embeddings for all claims in the corpus be $\mathcal{V}_C = \{\mathbf{v}(c) \mid c \in \bigcup_i C(x_i)\}$. For each claim $c_k$, local density is calculated using the $K$-nearest neighbor method. First, the $K$ nearest neighbors of $\mathbf{v}(c_k)$ are found, and local density is defined as the average cosine similarity with them:
\begin{equation}
    \text{density}(\mathbf{v}(c_k)) = \frac{1}{K} \sum_{j=1}^{K} \text{sim}(\mathbf{v}(c_k), \mathbf{v}^{(j)})
    \label{eq:density}
\end{equation}
Here, $\mathbf{v}^{(j)}$ is the $j$-th nearest neighbor of $\mathbf{v}(c_k)$, and $\text{sim}(\cdot, \cdot)$ is cosine similarity:
\begin{equation}
    \text{sim}(\mathbf{u}, \mathbf{v}) = \frac{\mathbf{u} \cdot \mathbf{v}}{\|\mathbf{u}\| \|\mathbf{v}\|}
    \label{eq:cosine}
\end{equation}
The claim rarity score is defined as the complement of density:
\begin{equation}
    \text{Rar}_C(c_k) = 1 - \text{density}(\mathbf{v}(c_k))
    \label{eq:rar_c}
\end{equation}
This design results in low scores when there are many similar claims nearby and high scores when there are few similar claims.

The claim rarity score for essay $x$ is calculated as the average rarity of all claims contained in that essay:
\begin{equation}
    S_{\text{claim}}(x) = \frac{1}{|C(x)|} \sum_{c_k \in C(x)} \text{Rar}_C(c_k)
    \label{eq:s_claim}
\end{equation}
When no claims exist ($|C(x)| = 0$), $S_{\text{claim}}(x) = 0$. This study used $K = 5$.

\subsubsection{Evidence Rarity}
Evidence rarity is calculated using the same method as claims. Let the set of embeddings for all evidence in the corpus be $\mathcal{V}_E = \{\mathbf{v}(e) \mid e \in \bigcup_i E(x_i)\}$, and define the rarity of each evidence $e_k$ as follows:
\begin{equation}
    \text{Rar}_E(e_k) = 1 - \text{density}(\mathbf{v}(e_k))
    \label{eq:rar_e}
\end{equation}
The evidence rarity score for essay $x$ is calculated as the average rarity of all evidence:
\begin{equation}
    S_{\text{evid}}(x) = \frac{1}{|E(x)|} \sum_{e_k \in E(x)} \text{Rar}_E(e_k)
    \label{eq:s_evid}
\end{equation}
When no evidence exists ($|E(x)| = 0$), $S_{\text{evid}}(x) = 0$.

\subsection{Cognitive Depth Assessment}
\label{subsec:cognitive_depth}

The cognitive depth score $S_{\text{move}}$ evaluates the depth of reasoning and completeness of response to counterarguments. While rarity assessments measure whether something is rare, cognitive depth evaluates how deeply arguments are developed.

The cognitive depth score consists of three components:
\begin{equation}
    S_{\text{move}}(x) = \alpha \cdot n_{\text{matched}} + \beta \cdot \text{Depth}(x) + \gamma \cdot \text{Coverage}(x)
    \label{eq:s_move}
\end{equation}
Here, $\alpha, \beta, \gamma$ are weight parameters for each component. These three components capture different aspects of critical thinking: $n_{\text{matched}}$ evaluates the quantity of critical dialogue, $\text{Depth}(x)$ evaluates the hierarchical complexity of argumentation, and $\text{Coverage}(x)$ evaluates the completeness of critical dialogue. Each component is explained below.

The first component, $n_{\text{matched}}$, is the number of counterargument-rebuttal matched pairs as defined in Section 3.2. The second component, $\text{Depth}(x)$, is the depth of reasoning as defined in Section 3.2. The third component, $\text{Coverage}(x)$, is the counterargument coverage rate, representing the proportion of presented counterarguments that are responded to with rebuttals:
\begin{equation}
    \text{Coverage}(x) = \begin{cases}
        \frac{n_{\text{matched}}}{n_A} & \text{if } n_A > 0 \\
        0 & \text{otherwise}
    \end{cases}
    \label{eq:coverage}
\end{equation}
This metric captures whether counterarguments are not only raised but also appropriately responded to. This study used $\alpha = \beta = \gamma = 1.0$, treating the three components equally. These weights can be adjusted according to educational objectives.

\subsection{Quality Adjustment}
\label{subsec:quality}

The quality score $Q(x)$ prevents essays that are original but logically flawed from receiving high scores. The quality score consists of two components: Coherence and Logical Validity:
\begin{equation}
    Q(x) = w_{\text{coh}} \cdot Q_{\text{coherence}}(x) + w_{\text{log}} \cdot Q_{\text{logical}}(x)
    \label{eq:quality}
\end{equation}
Here, $w_{\text{coh}}$ and $w_{\text{log}}$ are the weights for each component. This study used $w_{\text{coh}} = w_{\text{log}} = 0.5$.

This framework intentionally excludes fluency from quality evaluation, as LLM-generated text generally has high fluency, and including it would favor AI-generated text. Each component is evaluated using an LLM. For coherence evaluation, judgment is made based on three criteria: i) whether there are contradictions between claims and reasons, ii) whether there are contradictions between reasons and evidence, and iii) whether there are contradictions among multiple claims. Output is in three levels: 1.0 (no contradictions), 0.5 (partial contradictions), and 0.0 (serious contradictions). For logical validity evaluation, judgment is made based on three criteria: i) whether reasons logically support claims, ii) whether evidence appropriately backs up reasons, and iii) whether there are logical leaps or fallacies. Output is in three levels: 1.0 (valid), 0.5 (partial problems), and 0.0 (serious defects).

In both evaluations, the prompt explicitly instructs not to evaluate fluency or style, focusing evaluation on logical structure only.

\subsection{Final Score Computation}
\label{subsec:final_score}

The four sub-scores ($S_{\text{struct}}$, $S_{\text{claim}}$, $S_{\text{evid}}$, $S_{\text{move}}$) have different scales, so each is z-score normalized within the corpus:
\begin{equation}
    \tilde{S}_{\text{comp}}(x) = \frac{S_{\text{comp}}(x) - \mu_{\text{comp}}}{\sigma_{\text{comp}}}
    \label{eq:zscore}
\end{equation}
Here, $\mu_{\text{comp}}$ and $\sigma_{\text{comp}}$ are the corpus-wide mean and standard deviation of each component.

The normalized sub-scores are integrated through weighted linear combination:
\begin{equation}
    \tilde{S}_{\text{orig}}(x) = \lambda_1 \tilde{S}_{\text{struct}}(x) + \lambda_2 \tilde{S}_{\text{claim}}(x) + \lambda_3 \tilde{S}_{\text{evid}}(x) + \lambda_4 \tilde{S}_{\text{move}}(x)
    \label{eq:s_orig}
\end{equation}
Here, $\lambda_1, \lambda_2, \lambda_3, \lambda_4$ are weights satisfying $\sum_{i=1}^{4} \lambda_i = 1$. This study used equal weights $\lambda_i = 0.25$.

The final score is calculated by multiplying the normalized originality score by the quality score:
\begin{equation}
    S_{\text{final}}(x) = \tilde{S}_{\text{orig}}(x) \times Q(x)
    \label{eq:s_final}
\end{equation}
This multiplicative design ensures that logically flawed essays ($Q(x) \approx 0$) receive low final scores regardless of originality, positioning quality as a prerequisite for originality. Algorithm~\ref{alg:aroa} summarizes the overall processing procedure of AROA.

\begin{algorithm}[htbp]
\caption{AROA: Argument Rarity-based Originality Assessment}
\label{alg:aroa}
\begin{algorithmic}[1]
\Require essay collection $\mathcal{D} = \{x_i\}_{i=1}^N$, weight parameters $\lambda_1, \lambda_2, \lambda_3, \lambda_4, \alpha, \beta, \gamma, w_{\text{coh}}, w_{\text{log}}$
\Ensure Final score for each essay $\{S_{\text{final}}(x_i)\}_{i=1}^N$

\Statex \textbf{// Stage 1: Argument Structure Extraction}
\For{each $x_i \in \mathcal{D}$}
    \State $\mathcal{S}(x_i) \gets \text{LLM\_Extract}(x_i)$ \Comment{Extract 5 elements and relationships}
    \State $\text{Depth}(x_i), n_{\text{matched}}(x_i) \gets \text{ComputeFeatures}(\mathcal{S}(x_i))$
\EndFor

\Statex \textbf{// Stage 2: Embedding}
\For{each element $e \in \bigcup_i \mathcal{S}(x_i)$}
    \State $\mathbf{v}(e) \gets \text{SentenceBERT}(e)$
\EndFor

\Statex \textbf{// Stage 3: Originality Sub-scores}
\State $\{S_{\text{struct}}(x_i)\} \gets \text{StructuralRarity}(\{\boldsymbol{\phi}(x_i)\})$ \Comment{KDE-based}
\State $\{S_{\text{claim}}(x_i)\} \gets \text{SemanticRarity}(\{\mathbf{v}(c) \mid c \in C(x_i)\})$ \Comment{$K$-NN based}
\State $\{S_{\text{evid}}(x_i)\} \gets \text{SemanticRarity}(\{\mathbf{v}(e) \mid e \in E(x_i)\})$
\For{each $x_i \in \mathcal{D}$}
    \If{$n_A(x_i) > 0$}
        \State $\text{Coverage}(x_i) \gets n_{\text{matched}}(x_i) / n_A(x_i)$
    \Else
        \State $\text{Coverage}(x_i) \gets 0$
    \EndIf
    \State $S_{\text{move}}(x_i) \gets \alpha \cdot n_{\text{matched}}(x_i) + \beta \cdot \text{Depth}(x_i) + \gamma \cdot \text{Coverage}(x_i)$
\EndFor

\Statex \textbf{// Stage 4: Quality Adjustment}
\For{each $x_i \in \mathcal{D}$}
    \State $Q_{\text{coherence}}(x_i) \gets \text{LLM\_Evaluate}(x_i, \text{``coherence''})$
    \State $Q_{\text{logical}}(x_i) \gets \text{LLM\_Evaluate}(x_i, \text{``logical''})$
    \State $Q(x_i) \gets w_{\text{coh}} \cdot Q_{\text{coherence}}(x_i) + w_{\text{log}} \cdot Q_{\text{logical}}(x_i)$
\EndFor

\Statex \textbf{// Stage 5: Final Score Computation}
\For{each component $\text{comp} \in \{\text{struct}, \text{claim}, \text{evid}, \text{move}\}$}
    \State $\mu_{\text{comp}}, \sigma_{\text{comp}} \gets \text{Mean}(\{S_{\text{comp}}(x_i)\}), \text{Std}(\{S_{\text{comp}}(x_i)\})$
    \State $\tilde{S}_{\text{comp}}(x_i) \gets (S_{\text{comp}}(x_i) - \mu_{\text{comp}}) / \sigma_{\text{comp}}$ for all $i$
\EndFor
\For{each $x_i \in \mathcal{D}$}
    \State $\tilde{S}_{\text{orig}}(x_i) \gets \sum_{j=1}^{4} \lambda_j \tilde{S}_j(x_i)$
    \State $S_{\text{final}}(x_i) \gets \tilde{S}_{\text{orig}}(x_i) \times Q(x_i)$
\EndFor

\State \Return $\{S_{\text{final}}(x_i)\}_{i=1}^N$
\end{algorithmic}
\end{algorithm}

%% file: Authors/documents/4_experiment-and-analysis.tex
\section{Experiments and Analysis}
\label{sec:experiments}

\subsection{Dataset}
\label{subsec:setup}

This study used the AI Detection for Essays Dataset (AIDE)~\citep{34}, a publicly available dataset on Kaggle constructed by Vanderbilt University and The Learning Agency Lab for AI-generated text detection in educational settings. AIDE contains argumentative essays written by middle and high school students alongside essays generated by multiple LLMs. The essays address two topics---Car-free cities and Electoral College---both requiring students to read given source materials and argue their own position. This dataset was selected because it contains numerous human essays on the same topic, suitable for rarity-based evaluation, and includes both human and AI essays for comparative analysis.

Table~\ref{tab:dataset} shows the dataset statistics. Human essays were collected from student submissions. The original AIDE dataset contains AI essays generated using multiple LLMs including PaLM 2 (text-bison, text-unicorn), Gemini 1.0 (Pro, Ultra), and GPT-4, with 3-shot prompting using randomly selected human essays as in-context examples and random sampling at temperature 1.0--1.6. Post-processing included deduplication (85\% cosine similarity threshold), removal of implausible outputs, formatting normalization, and statistical matching of text characteristics to human essays; approximately 10\% of originally generated essays were retained. Since the original AIDE dataset contained fewer AI essays than needed for balanced comparison, we supplemented the AI essay collection to reach 1,000 essays (500 per topic) by generating additional essays using GPT-4.1-mini following the same generation protocol as AIDE: 3-shot prompting with randomly selected human essays as in-context examples, the same assignment instructions and source materials, and equivalent post-processing steps.

\begin{table}[t]
    \centering
    \caption{Dataset statistics}
    \label{tab:dataset}
    \begin{tabular}{lccc}
        \hline
        Category & Car-free cities & Electoral College & Total \\
        \hline
        Human essays & 707 & 668 & 1,375 \\
        AI essays & 500 & 500 & 1,000 \\
        \hline
        Total & 1,207 & 1,168 & 2,375 \\
        \hline
    \end{tabular}
\end{table}

For the argument structure extraction and quality evaluation in the AROA pipeline, this study used GPT-4.1-mini~\citep{35}, which offers high instruction-following capability at low API cost, making it suitable for processing large-scale datasets. Other parameters follow the settings described in Section 3.

\subsection{Evaluation on Originality Score Distribution}
\label{subsec:score_distribution}

\subsubsection{Overall Score Distribution}

\begin{figure}[t]
    \centering
    \includegraphics[width=0.8\linewidth]{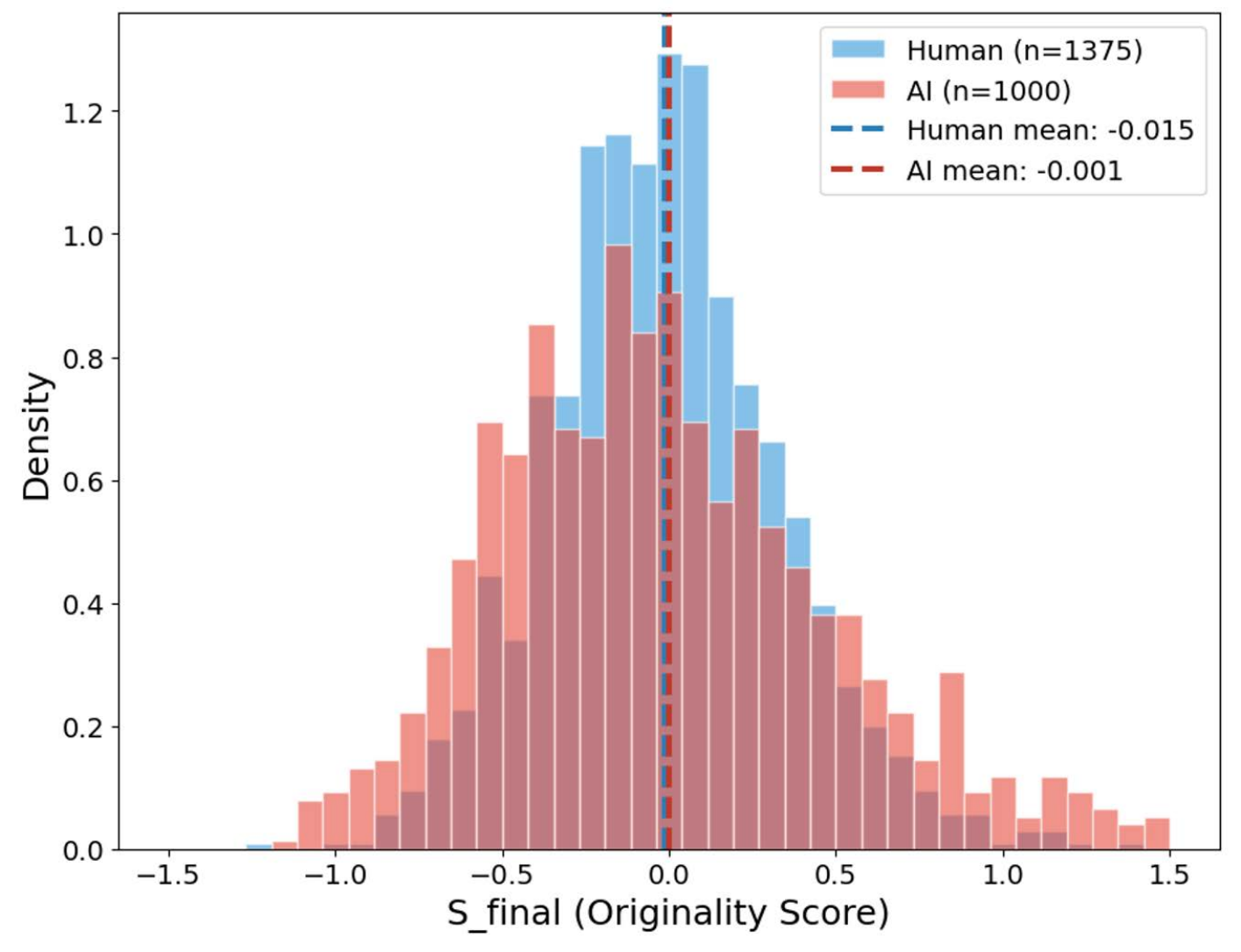}
    \caption{Distribution of final originality scores ($S_{\text{final}}$) for human essays and AI essays. Dashed lines indicate the mean of each group.}
    \label{fig:distribution}
\end{figure}

Figure~\ref{fig:distribution} shows the distribution of final originality scores $S_{\text{final}}$ for human essays ($n=1,375$) and AI essays ($n=1,000$). Both distributions overlap around the mean (human: $-0.015$, AI: $-0.001$), but differ in shape: human essays are concentrated in a narrow range (SD = 0.345), while AI essays are more dispersed (SD = 0.534), suggesting that AI originality varies with prompting and sampling conditions.

Table~\ref{tab:comparison} summarizes the statistical comparison. The final originality score $S_{\text{final}}$ showed no statistically significant difference between human essays and AI essays ($p = 0.118$, Cohen's $d = -0.032$). This indicates that AROA evaluates the originality of argumentation itself rather than functioning as an AI detector: human essays receive low scores if argumentation is typical, and AI essays receive high scores if argumentation is original.

\begin{table}[t]
    \centering
    \caption{Statistical comparison between human and AI essays}
    \label{tab:comparison}
    \begin{tabular}{lcccc}
        \hline
        Metric & Human ($n=1,375$) & AI ($n=1,000$) & Cohen's $d$ & $p$-value \\
        \hline
        $S_{\text{final}}$ & $-0.015 \pm 0.345$ & $-0.001 \pm 0.534$ & $-0.032$ & 0.118 \\
        $S_{\text{struct}}$ & $4.987 \pm 0.962$ & $4.797 \pm 0.885$ & 0.204 & $<0.001$*** \\
        $S_{\text{claim}}$ & $0.170 \pm 0.076$ & $0.037 \pm 0.025$ & 2.239 & $<0.001$*** \\
        $S_{\text{evid}}$ & $0.092 \pm 0.042$ & $0.048 \pm 0.014$ & 1.343 & $<0.001$*** \\
        $S_{\text{move}}$ & $5.093 \pm 2.159$ & $5.156 \pm 2.340$ & $-0.028$ & 0.821 \\
        $Q$ & $0.609 \pm 0.141$ & $0.998 \pm 0.021$ & $-3.600$ & $<0.001$*** \\
        \hline
    \end{tabular}
\end{table}

\subsubsection{Component-wise Analysis}

\begin{figure}[t]
    \centering
    \includegraphics[width=\linewidth]{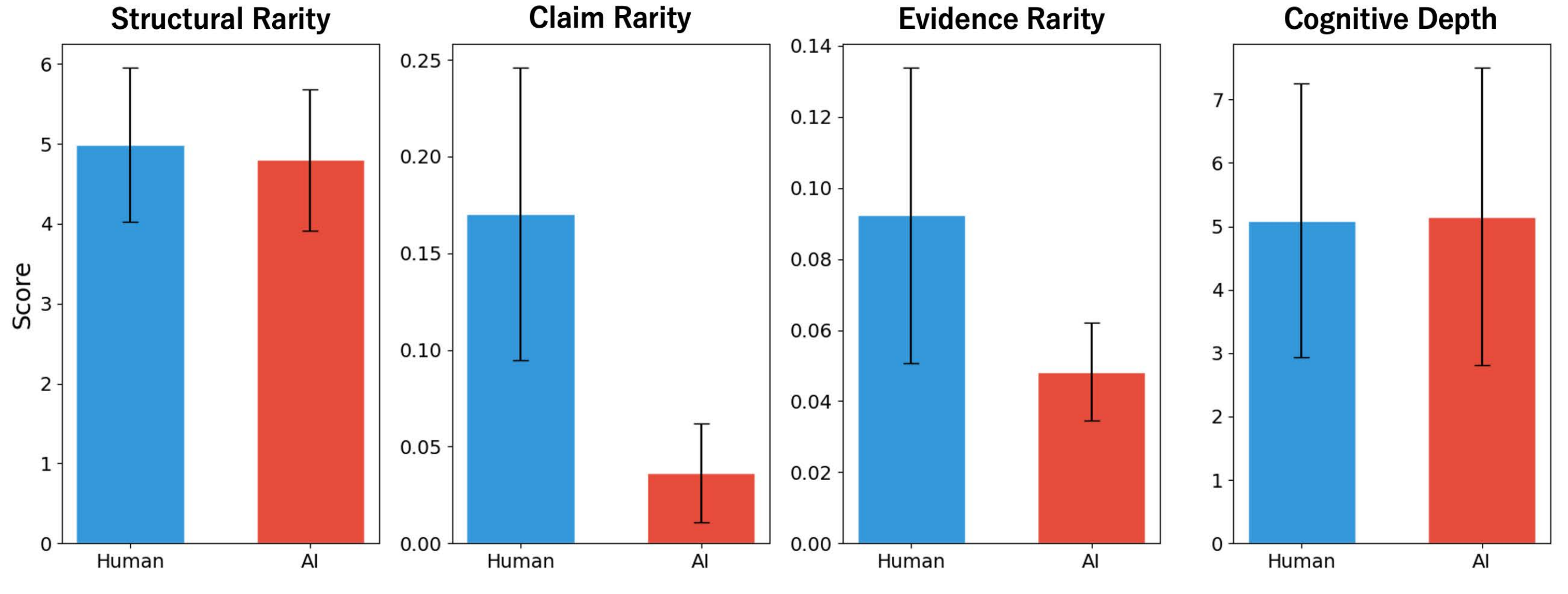}
    \caption{Comparison of four originality components between human and AI essays: Structural Rarity ($S_{\text{struct}}$), Claim Rarity ($S_{\text{claim}}$), Evidence Rarity ($S_{\text{evid}}$), and Cognitive Depth ($S_{\text{move}}$).}
    \label{fig:components}
\end{figure}

Figure~\ref{fig:components} shows the comparison between human essays and AI essays for each of the four originality components. Although overall scores are similar, both groups showed markedly different component profiles.

In claim rarity $S_{\text{claim}}$, human essays scored substantially higher (human: $0.170$, AI: $0.037$, $d = 2.24$), with AI scores at approximately one-fifth of human levels. AI-generated essays tend to converge on typical claim patterns from training data, resulting in a clustered distribution in semantic space.

A similar trend was observed in evidence rarity $S_{\text{evid}}$ (human: $0.092$, AI: $0.048$, $d = 1.34$), with human essays showing more diverse evidence selection while AI essays relied on commonly cited evidence.

In contrast, no substantial differences were observed in structural rarity $S_{\text{struct}}$ ($d = 0.20$) and cognitive depth $S_{\text{move}}$ ($d = -0.03$), indicating that AI can reproduce the structural complexity of human argumentation.

\subsection{Evaluation on Quality-Originality Relationship}
\label{subsec:quality_originality}

\subsubsection{Component Independence}

\begin{figure}[t]
    \centering
    \includegraphics[width=0.7\linewidth]{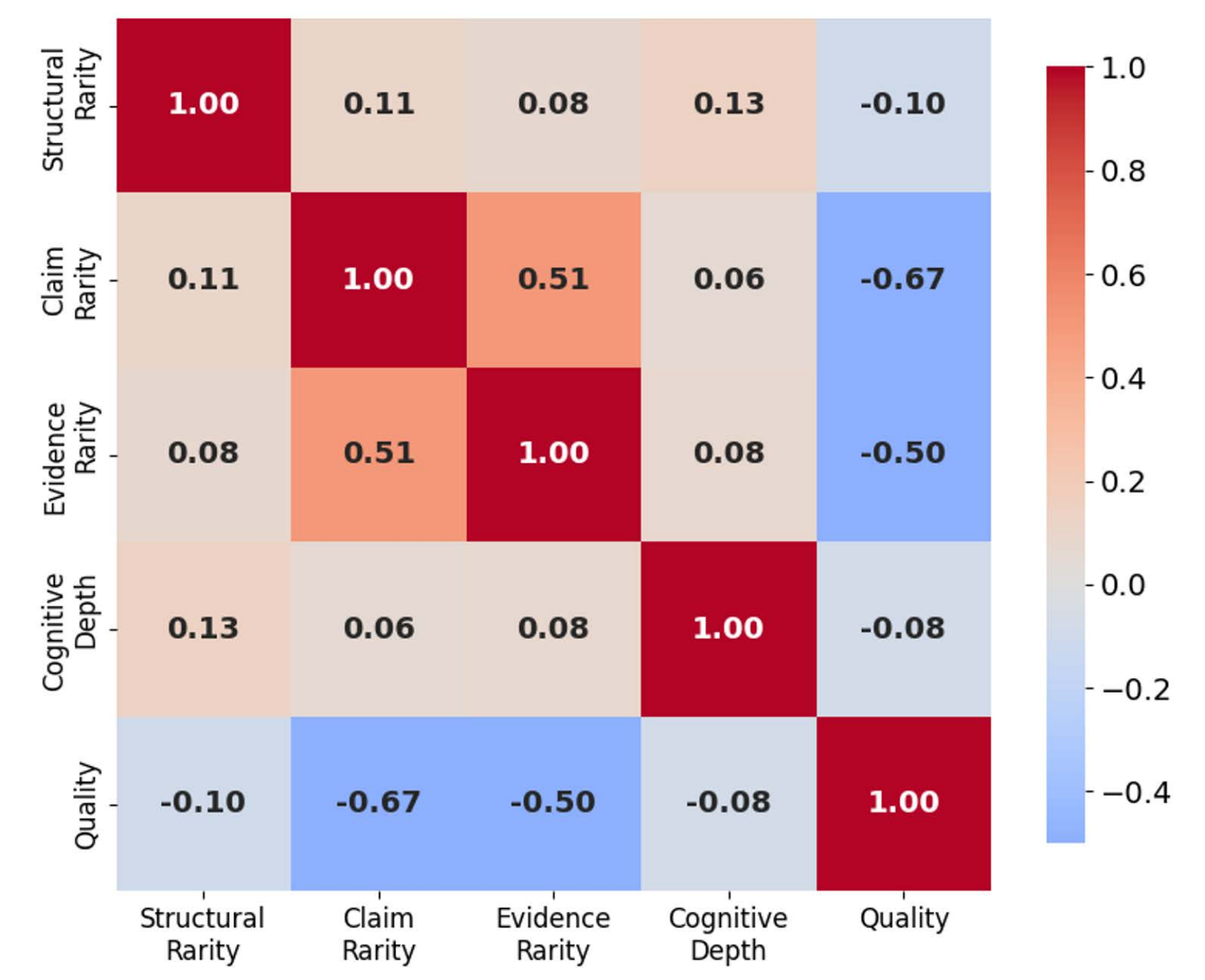}
    \caption{Correlation matrix of originality components and quality score.}
    \label{fig:correlation}
\end{figure}

Figure~\ref{fig:correlation} shows the correlation matrix among originality components and quality score. First, structural rarity $S_{\text{struct}}$ has extremely low correlations with all other components ($r = 0.08$--$0.13$). This indicates that structural atypicality is independent of claim/evidence content and depth of reasoning, supporting the design philosophy that how arguments are made and what is argued are different evaluation axes.

Cognitive depth $S_{\text{move}}$ also has low correlations with other components ($r = 0.06$--$0.13$), confirming that reasoning depth varies independently of semantic rarity.

A moderate positive correlation ($r = 0.51$) between $S_{\text{claim}}$ and $S_{\text{evid}}$ suggests that authors with original claims tend to also present unique evidence, though the components evaluate different argument elements and do not fully coincide.

These results support the validity of multifaceted evaluation, as each component captures different aspects of originality.

\subsubsection{Quality Does Not Guarantee Originality}

The most striking finding in the correlation matrix is the strong negative correlation between quality score $Q$ and semantic rarity components. The correlation between $Q$ and $S_{\text{claim}}$ is $r = -0.67$, and between $Q$ and $S_{\text{evid}}$ is $r = -0.50$, both showing statistically significant strong negative correlations. This result means that higher-quality essays tend to have lower rarity of claims and evidence.

This reflects a quality-rarity trade-off: logically coherent, high-quality text is easier to produce using widely accepted claim patterns, while original claims and rare evidence make maintaining coherence more difficult.

On the other hand, correlations between $Q$ and structural components ($S_{\text{struct}}$: $r = -0.10$, $S_{\text{move}}$: $r = -0.08$) are extremely weak. This indicates that the complexity and atypicality of argument structure do not significantly affect quality. Even essays with complex structures can maintain logical coherence.

\begin{figure}[t]
    \centering
    \includegraphics[width=0.7\linewidth]{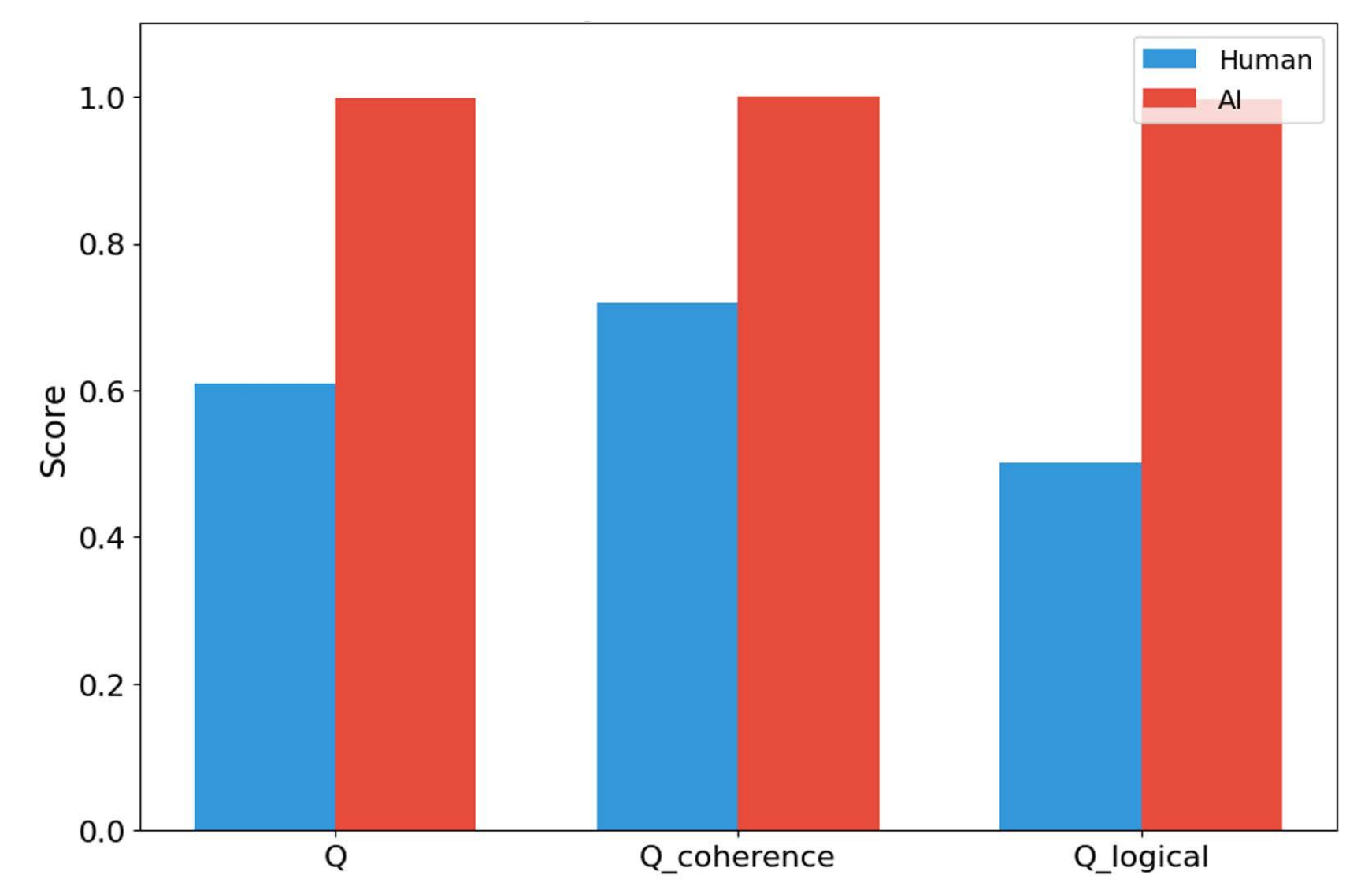}
    \caption{Comparison of quality scores between human and AI essays.}
    \label{fig:quality}
\end{figure}

Figure~\ref{fig:quality} shows the human-AI comparison of quality score $Q$ and its components ($Q_{\text{coherence}}$, $Q_{\text{logical}}$). AI essays substantially exceeded human essays in all quality metrics. In particular, AI essay quality scores are nearly perfect ($Q = 0.998$), achieving the highest levels in both coherence ($Q_{\text{coherence}} \approx 1.0$) and logical validity ($Q_{\text{logical}} \approx 1.0$). On the other hand, human essays remain at $Q = 0.609$, with room for improvement particularly in logical validity ($Q_{\text{logical}} \approx 0.50$).

These results demonstrate that text quality does not guarantee originality. AI essays achieve high quality by relying on typical claim patterns, while human essays show higher semantic originality but lower quality scores. The multiplicative design (Equation~\ref{eq:s_final}) ensures that low originality in AI essays is partially compensated by high $Q$, while high originality in human essays is adjusted by variation in $Q$, resulting in similar final score distributions and confirming that quality and originality are independent axes.

\subsubsection{Illustrative Examples}

To concretely illustrate the relationship between quality and originality, representative essays were extracted from the AIDE dataset and their scores and content were compared. Table~\ref{tab:illustrative} shows the scores of high-originality and low-originality essays.

\begin{table}[t]
    \centering
    \caption{Comparison of high-originality and low-originality essays}
    \label{tab:illustrative}
    \begin{tabular}{lcccccc}
        \hline
        Essay & $S_{\text{final}}$ & $S_{\text{struct}}$ & $S_{\text{claim}}$ & $S_{\text{evid}}$ & $S_{\text{move}}$ & $Q$ \\
        \hline
        High-originality & 1.410 & 7.444 & 0.294 & 0.135 & 10.0 & 0.75 \\
        Low-originality & $-1.230$ & 4.100 & 0.038 & 0.046 & 3.0 & 1.00 \\
        \hline
    \end{tabular}
\end{table}

The high-originality essay ($S_{\text{final}} = 1.41$) argues from multiple perspectives on the topic of Car-free cities. The following is an excerpt:

\begin{quote}
\textit{``having a society like this is also very stress relieving in a sense. you no longer have the every day stress of being on a highway or constantly alert for cars [...] the amount of exercise is also increased here because you are walking or biking everywhere''}
\end{quote}

\begin{quote}
\textit{``Now with all this being said there obviously are still downsides to cutting out gas producing cars. weather being a main one. suppose it rains one day, how do you get to work? [...] if you work out of town how do you get to your job?''}
\end{quote}

This essay received a high score because it presents original perspectives beyond environmental issues ($S_{\text{claim}} = 0.294$), raises and responds to counterarguments ($S_{\text{move}} = 10.0$), and its original content compensates for imperfect quality ($Q = 0.75$).

On the other hand, the low-originality essay ($S_{\text{final}} = -1.23$) takes a typical approach to the same topic:

\begin{quote}
\textit{``The actual amount of people driving and getting their license has decreased over the past couple years and this is due to two reasons, not driving helps the environment and it helps the community.''}
\end{quote}

\begin{quote}
\textit{``There has been large amounts of information drawn from certain `Car-Free' experiments around the world. It has shown that the limited usage of cars has improved the environment by reducing smog in Paris and has also helped communities like Vauban increase the happiness of its citizens.''}
\end{quote}

Despite perfect quality ($Q = 1.00$), this essay scores low because it relies on stereotypical claims ($S_{\text{claim}} = 0.038$), lists citations without original analysis ($S_{\text{evid}} = 0.046$), and lacks counterarguments ($S_{\text{move}} = 3.0$).

These examples illustrate the distinction between conventional quality-focused assessment and AROA's originality-focused evaluation.

\subsection{Evaluation on Robustness to Sample Size}
\label{subsec:robustness}

Since the proposed method is based on density estimation, the reference corpus size may affect score reliability. Subsampling experiments were conducted for sample sizes $n \in \{30, 50, 100, 200, 300, 500, 700, 1000\}$, repeated 5 times each. For each subsample, $S_{\text{struct}}$ (the component most sensitive to sample size) was recalculated and Spearman's $\rho$ was computed against full-data rankings.

\begin{figure}[t]
    \centering
    \includegraphics[width=0.8\linewidth]{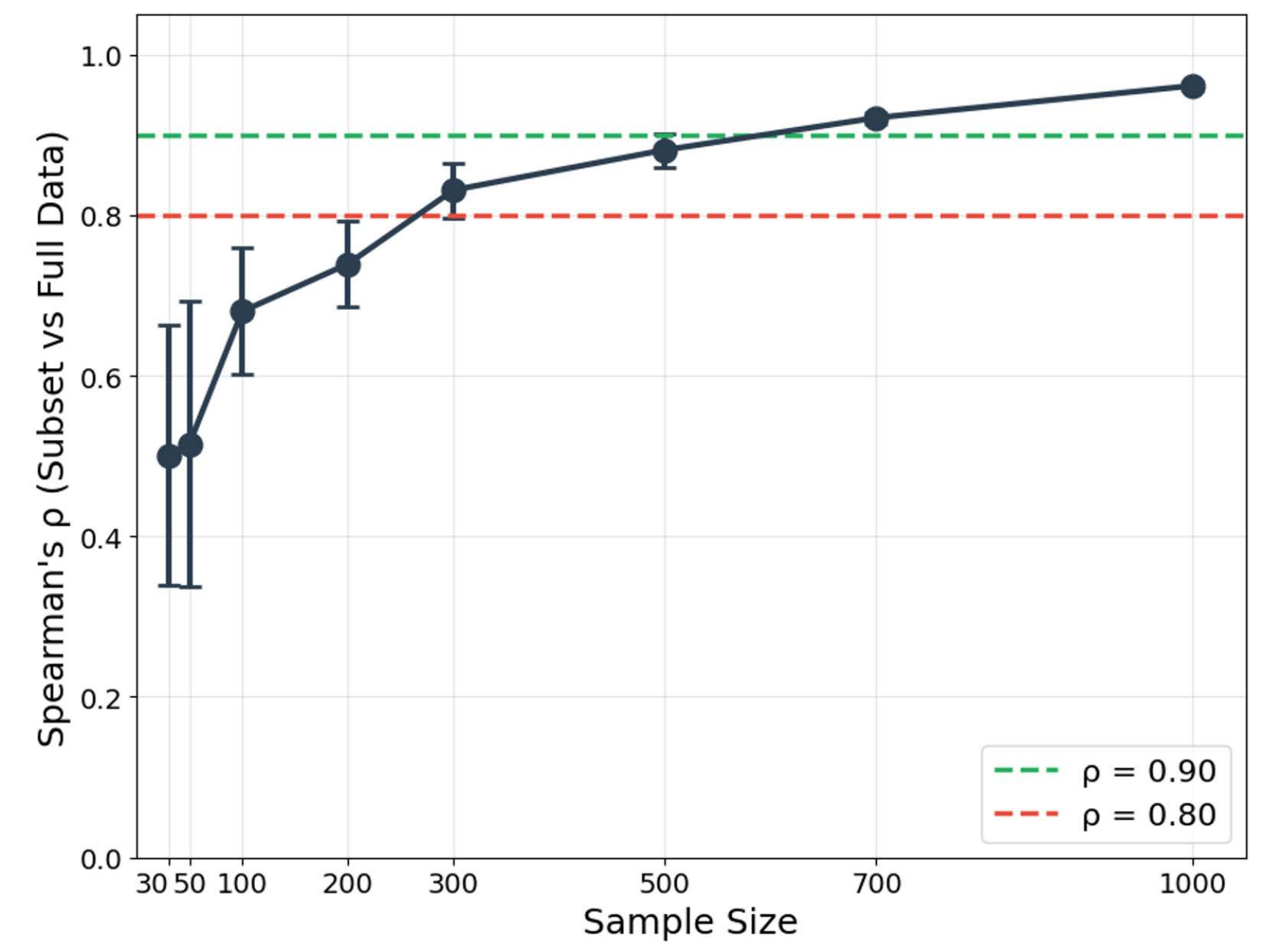}
    \caption{Rank correlation stability against sample size. The plot shows Spearman's rank correlation ($\rho$) between subset-based and full-data-based rankings at each sample size. Error bars indicate standard deviation across 5 trials.}
    \label{fig:robustness}
\end{figure}

Figure~\ref{fig:robustness} shows the relationship between sample size and rank correlation. Rank correlation improves monotonically with increasing sample size, exceeding the $\rho = 0.80$ threshold around $n=300$ and the $\rho = 0.90$ threshold around $n=700$. As the error bars indicate, variation between trials also decreases as sample size increases. On the other hand, at typical classroom scales ($n = 30$--$50$), rank correlation remains at approximately 0.50. These results indicate that when actually applying the proposed method, the reference corpus size needs to be considered.

\subsection{Cost Analysis}
\label{subsec:cost}

The experiments were conducted on a macOS environment equipped with an Apple M1 chip (8-core CPU, 7-core GPU) and 16GB RAM, using 10 parallel workers. Table~\ref{tab:cost_breakdown} shows the processing time breakdown per essay. Approximately 90\% of processing time is occupied by argument extraction through LLM API calls, while embedding and density estimation are negligible as local processing.

\begin{table}[t]
    \centering
    \caption{Processing time breakdown per essay}
    \label{tab:cost_breakdown}
    \begin{tabular}{lrl}
        \hline
        Component & Time & Note \\
        \hline
        Argument extraction & 1.35 sec & GPT-4.1-mini API \\
        Text embedding & 0.04 sec & Local Sentence-BERT \\
        Density estimation & $<$0.01 sec & NumPy/scikit-learn \\
        Quality evaluation & 0.13 sec & GPT-4.1-mini API \\
        \hline
        Total & 1.52 sec & \\
        \hline
    \end{tabular}
\end{table}

Table~\ref{tab:cost_scalability} shows measured values from this experiment and estimated costs assuming application at various scales. The measured time for processing 1,375 human essays was 34.8 minutes with an API cost of \$3.36, approximately \$0.0024 (about 0.24 cents) per essay. From classroom scale (30 essays) to institution scale (10,000 essays), processing time and cost scale linearly, enabling operation at realistic costs even for large-scale applications.

\begin{table}[t]
    \centering
    \caption{Scalability analysis: processing time and API cost}
    \label{tab:cost_scalability}
    \begin{tabular}{rrrl}
        \hline
        Scale & Essays & Time & API Cost \\
        \hline
        Single class & 30 & 0.8 min & \$0.07 \\
        Multiple classes & 100 & 2.5 min & \$0.24 \\
        Course level & 500 & 13 min & \$1.20 \\
        Department level & 1,375 & 35 min & \$3.36 (measured) \\
        Large-scale & 5,000 & 2.1 hours & \$12.00 \\
        Institution level & 10,000 & 4.2 hours & \$24.00 \\
        \hline
    \end{tabular}
\end{table}

Compared to human evaluation (5--10 minutes and \$1--5 per essay), the proposed method is 200--400 times faster (1.5 seconds) and 400--2,000 times cheaper (\$0.0024), enabling large-scale originality assessment that was previously impractical.

\subsection{Evaluation on Transferability Across LLMs}
\label{subsec:transferability}

To verify whether the proposed framework depends on the specific LLM used, transferability analysis was conducted using GPT-4.1-mini, Gemini 2.5 Flash Lite, and Claude 3 Haiku.

The average cross-model Pearson correlation for $S_{\text{final}}$ was $r = 0.465$ (Table~\ref{tab:transfer_overall}), confirming moderate agreement. GPT showed relatively high agreement with both Gemini and Claude, while Gemini--Claude correlation was lowest ($r = 0.402$).

\begin{table}[t]
    \centering
    \caption{Cross-model correlation of $S_{\text{final}}$}
    \label{tab:transfer_overall}
    \begin{tabular}{lcc}
        \hline
        Model pair & Pearson $r$ & Spearman $\rho$ \\
        \hline
        GPT--Gemini & 0.505 & 0.501 \\
        GPT--Claude & 0.489 & 0.496 \\
        Gemini--Claude & 0.402 & 0.416 \\
        \hline
        Average & 0.465 & 0.471 \\
        \hline
    \end{tabular}
\end{table}

Component-wise analysis (Table~\ref{tab:transfer_components}) revealed that evidence rarity $S_{\text{evid}}$ showed the highest stability (average $r = 0.616$), while structural rarity $S_{\text{struct}}$ (average $r = 0.232$) and quality $Q$ (average $r = 0.249$) showed low transferability. Claude in particular assigned nearly uniform quality scores, suggesting fundamentally different evaluation criteria.

\begin{table}[t]
    \centering
    \caption{Cross-model correlation by component (Pearson $r$)}
    \label{tab:transfer_components}
    \begin{tabular}{lccccc}
        \hline
        Component & GPT--Gemini & GPT--Claude & Gemini--Claude & Average & Stability \\
        \hline
        $S_{\text{evid}}$ & 0.658 & 0.626 & 0.563 & 0.616 & High \\
        $S_{\text{claim}}$ & 0.462 & 0.459 & 0.532 & 0.484 & Medium \\
        $S_{\text{move}}$ & 0.475 & 0.460 & 0.352 & 0.429 & Medium \\
        $Q$ & 0.483 & 0.114 & 0.148 & 0.249 & Low \\
        $S_{\text{struct}}$ & 0.294 & 0.268 & 0.133 & 0.232 & Low \\
        \hline
    \end{tabular}
\end{table}

\begin{figure}[t]
    \centering
    \includegraphics[width=0.8\linewidth]{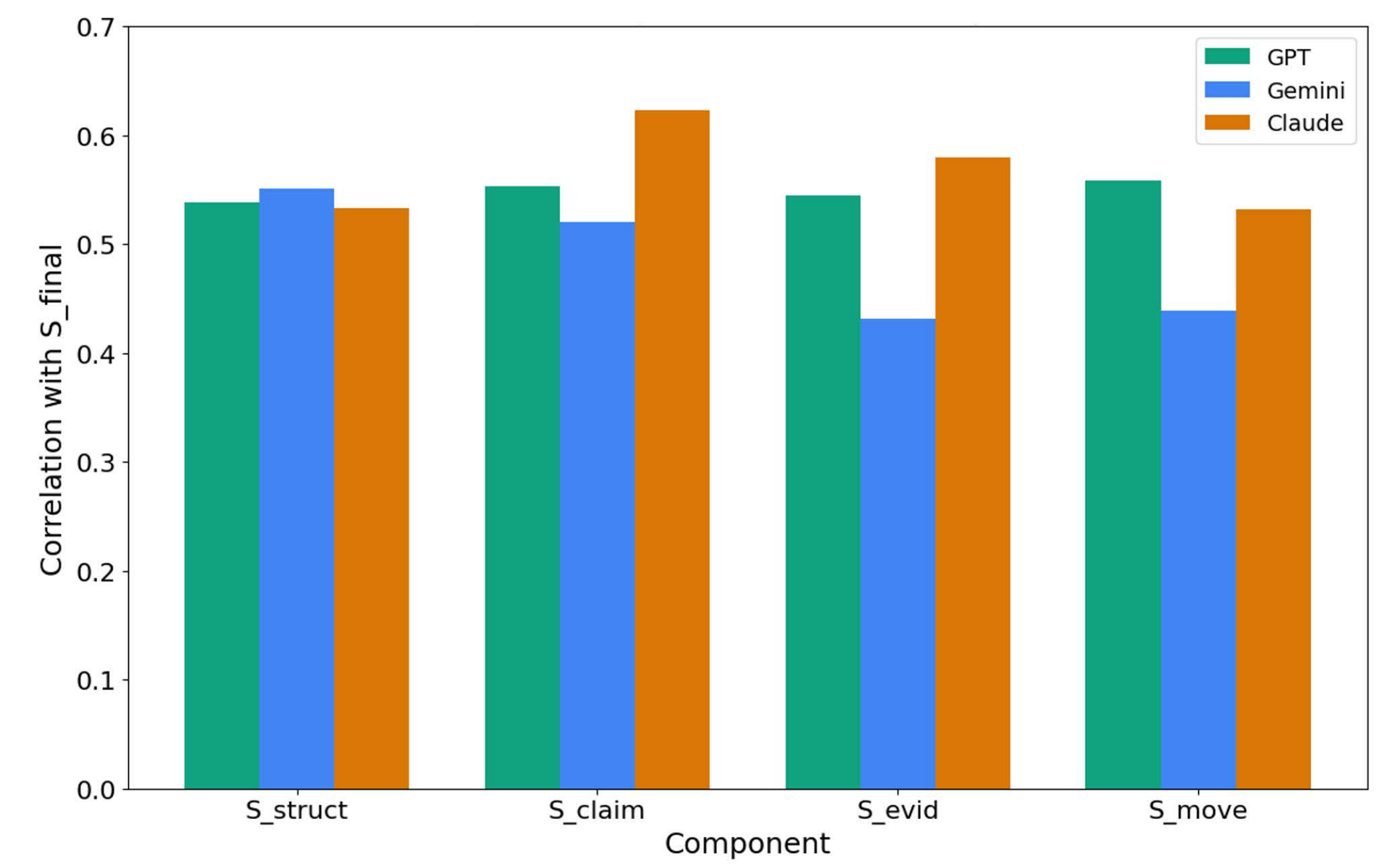}
    \caption{Component impact on final score by model.}
    \label{fig:component_impact}
\end{figure}

Each model showed characteristic evaluation tendencies (Figure~\ref{fig:component_impact}): GPT contributed equally across components ($r = 0.54$--$0.56$), Gemini emphasized structural rarity ($r = 0.55$), and Claude emphasized semantic rarity ($S_{\text{claim}}$: $r = 0.62$, $S_{\text{evid}}$: $r = 0.58$).

These results indicate that AROA captures argumentative originality regardless of the LLM adopted, though component stability varies: evidence scores transfer well, while structure and quality evaluations are more model-dependent.

%% file: Authors/documents/5_discussion.tex
\section{Discussion}
\label{sec:discussion}

\subsection{Key Findings}
\label{subsec:key_findings}

The experiments using 1,375 human essays and 1,000 AI essays yielded three key findings. First, a strong negative correlation between text quality and semantic rarity ($r = -0.67$ for $Q$ vs. $S_{\text{claim}}$; $r = -0.50$ for $Q$ vs. $S_{\text{evid}}$) reveals a quality-originality trade-off: logically coherent, high-quality text is easier to produce using widely accepted claims and common evidence, while original claims make maintaining coherence more difficult.

Second, human essays and AI essays exhibit markedly different component profiles despite similar final scores ($p = 0.118$). AI essays achieved near-perfect quality ($Q = 0.998$) but their claim rarity remained at approximately one-fifth of human levels (AI: 0.037, human: 0.170). In contrast, no significant differences were observed in structural rarity ($d = 0.20$) or cognitive depth ($d = -0.03$). This indicates that contemporary LLMs can reproduce the structural complexity of human argumentation but tend to converge on typical semantic patterns from training data.

Third, the four originality components capture genuinely different aspects of originality. Structural rarity and cognitive depth showed low correlations with other components ($r = 0.06$--$0.13$), while claim and evidence rarity showed a moderate correlation ($r = 0.51$), confirming that how arguments are made and what is argued are independent evaluation axes.

\subsection{Theoretical Implications}
\label{subsec:theoretical_implications}

These findings contribute to the theoretical understanding of writing assessment and AI-generated text in several ways. First, the quality-originality trade-off challenges the longstanding assumption in AES research that text quality is the primary evaluation target. Our results demonstrate that quality and originality are not only independent but inversely related in their semantic dimensions, suggesting that conventional quality-focused assessment systematically undervalues original thinking. This extends the creativity assessment literature~\citep{23, 24}, which has established rarity as a measure of originality for short responses in divergent thinking tests, to the domain of complex argumentative writing.

Second, the component analysis provides empirical evidence for the multidimensional nature of argumentative originality. The low correlations between structural and semantic components support the theoretical distinction between the form and content of argumentation, consistent with Toulmin's framework~\citep{17}. This finding implies that originality assessment frameworks must adopt multifaceted approaches rather than relying on single indicators.

Third, the human-AI comparison reveals that the limitations of current LLMs lie not in structural competence but in semantic originality. This finding contributes to the growing literature on AI capabilities and limitations by identifying a specific dimension, content originality, where human writers retain a clear advantage. Unlike AI detection approaches that focus on surface-level artifacts, AROA provides a content-level framework for understanding the qualitative differences between human and AI writing.

\subsection{Practical Implications}
\label{subsec:practical_implications}

From a practical perspective, AROA has sufficient efficiency for deployment in educational settings, with processing costs of approximately 1.5 seconds and \$0.0024 per essay. Robustness analysis showed that rank correlation $\rho \geq 0.80$ is achieved with a reference corpus of 300 or more essays, yielding reliable results for courses pooling multiple classes. The RaaS implementation at a large Japanese university, which accommodates up to 1,000 students per class~\citep{36}, exemplifies settings well-suited for density-based originality assessment: larger cohorts provide richer reference corpora, making the scale of large enrollment courses an advantage rather than a limitation. For small-scale classes (30--50 essays), where rank correlation remains at approximately 0.50, the framework is recommended for formative feedback rather than summative ranking.

Transferability analysis confirmed that AROA does not heavily depend on a specific LLM ($r = 0.465$ average cross-model correlation for $S_{\text{final}}$). Evidence rarity showed the highest stability ($r = 0.616$), while structural rarity and quality evaluation were more model-dependent ($r = 0.23$--$0.25$). Each model showed characteristic evaluation tendencies: GPT performed balanced evaluation, Gemini emphasized structure, and Claude emphasized semantic rarity, suggesting the possibility of more robust evaluation through model ensembles.

These results suggest that AROA can complement the limitations of human evaluation in assessing originality. Human evaluators face order effects and fatigue~\citep{5}, expertise bias~\citep{6}, and cognitive limitations in judging relative rarity across entire corpora. AROA addresses these by providing objective, consistent, corpus-wide rarity quantification, offering educators a foundation for more fairly evaluating and developing students' critical thinking skills.

\subsection{Limitations and Future Work}
\label{subsec:limitations}

This study has several limitations. First, the evaluation is based on a single dataset limited to two topics, and generalizability to different topics, languages, grade levels, and genres (e.g., academic writing, creative writing) requires future verification. Second, the framework has not been validated against originality evaluations by human experts; empirical research is needed to determine how well AROA scores align with expert judgments of originality. Third, since argument extraction and quality evaluation depend on LLMs, model biases may affect scores. Although transferability analysis confirmed moderate cross-model agreement, structural rarity and quality evaluation showed high model dependency. Fourth, reliable evaluation requires a reference corpus of 300 or more essays, constraining application in small-scale classes.

Future work should address these limitations through: i) verification of generalizability using diverse datasets, ii) empirical validation through correlation with human expert evaluation, iii) development of feedback generation functionality for formative assessment, and iv) longitudinal studies tracking originality development over time.

%% file: Authors/documents/6_conclusion.tex
\section{Conclusion}
\label{sec:conclusion}

This study proposed AROA, a framework for automatically evaluating argumentative originality in essays through four complementary components, structural rarity, claim rarity, evidence rarity, and cognitive depth, using density-based rarity quantification with quality adjustment. Experiments using 1,375 human essays and 1,000 AI essays revealed three key findings: i) a strong quality-originality trade-off ($r = -0.67$ between quality and claim rarity), ii) that contemporary LLMs reproduce structural complexity but exhibit claim rarity at approximately one-fifth of human levels, and iii) that the four components capture genuinely independent aspects of originality. These results demonstrate that quality and originality are independent evaluation axes and that assessment must shift accordingly in the AI era. Future work includes generalizability verification across diverse datasets, empirical validation with human expert evaluation, and development of feedback generation for formative assessment.